# Color-Coded Symbology and New Computer Vision Tool to Predict the Historical Color Pallets of the Renaissance Oil Artworks

Artyom M. Grigoryan*[1] and Sos S. Agaian [2]

[1] Department of Electrical and Computer Engineering,
The University of Texas at San Antonio, USA
`amgrigoryan@utsa.edu`

[2] City University of New York / CSI

## Abstract

In this paper, we discuss possible color palletes, prediction and analysis of originality of the colors that Artists used on the Renaissance oil paintings. This framework goal is to help to use the color symbology and image enhancement tools, to predict the historical color palletes of the Renaissance oil artworks. This work is only the start of a development to explore the possibilities of prediction of color palletes of the Renaissance oil artworks. We believe that framework might be very useful in the prediction of color palletes of the Renaissance oil artworks and other artworks. The images in number 105 have been taken from the paintings of three well-known artists, Rafael, Leonardo Da Vinci, and Rembrandt that are available in the Olga's Gallery by address http://www.abcgallery.com/. Images are processed in the frequency domain to enhance a quality of images and ratios of primary colors are calculated and analyzed by using new measurements of color-ratios.

## 1. Introduction

Recently, we have seen an enormous interest of the application of computer science, computer vision, color science, digital forensics, and pattern recognition tools to the problems in the priceless art/paintings [1]-[7]. Utmost all these studies are focused on image analysis, understanding, art conservation, dissemination, transmission, representation, and automatic visual examination and restoration to provide reliable information to distinguish artists, for other, color palette and others. We perhaps may identify the following investigation and application main areas: the achievement of a digital version of traditional photographic reproductions, image acquisition, archiving, automatically indexing images, Web-based image sharing, digital forensic analysis, art historical studies, image diagnostics, the painting conditions evaluation and virtual restoration. In fact, these artwork applications are still very uncommon practices, among both conservators and information, computer science, forensic, and communication technology specialists because each work of art is, by its nature, unique (styles, dimensions, materials, color rejuvenation, time periods and artists). Moreover, color-coded symbology has been used by Renaissance artists because for each color, there is both a psychological and physiological impact paintings [8]-[15].

In this paper, we consider the problem of prediction of initial colors of paints in the works of recognized artists. With the time, many paintings lost their original view, or shine of colors. In this paper, we try to analyze colors of the image, their meaning, color ratio, and predict their original state in the proposed model. The image first is processed by the method of alpha-rooting to improve the quality of the image and then the colors are analyzed, by using a few measures that calculate the quality of the image, color ratios, and then the images are modified, by keeping their ratios to be equal to a unique ratio. This ratio that we call the artist ratio is calculated from the image by using the model described below. The color images are considered in the RGB model, with three primary colors, red, green, and blue. We introduced new measures to analyze the red-to-blue, green-to-blue, and blue-to-red color relationships.

## 2. PROCESSING IMAGES OF PAINTINGS

In this section, we consider the model of image processing with the color enhancement that is considered as color prediction or restoration, when the images of paintings are processed first in the frequency domain and then in the spatial domain. We assume that the pictures in their original early form were more pronounced and many details of the pictures were clearly visible and over time some colors faded in places and the details were smoothed out. It is also possible that some unknown proportions of colors have also changed. Therefore, we will process the images in the frequency domain to enhance the visibility and quality of paintings and will analyze the ratio of colors, by assuming that each of famous artists had a unique color proportion in the paintings. Base on these assumptions, we propose the process of color images of painting that is described by the block-diagram given in Fig. 1.

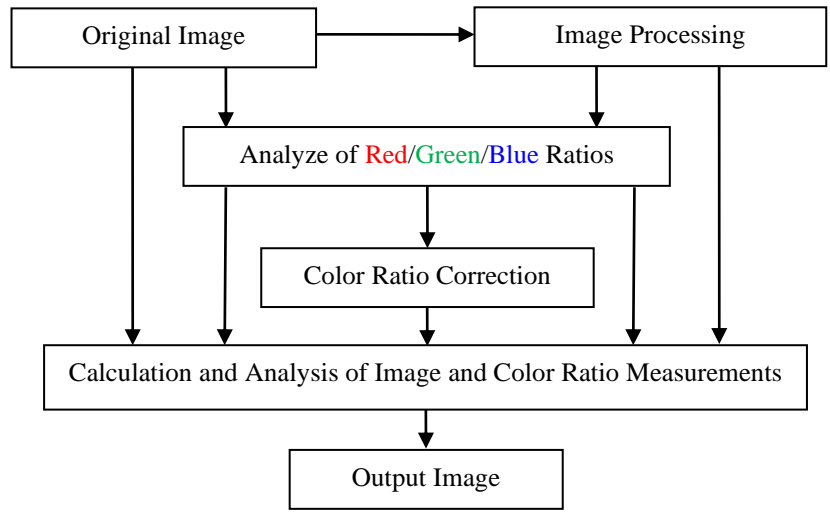

**Fig. 1**. Block-diagram of the color prediction/restoration in images of paintings.

The original image is processed in the frequency domain and then the color ratios are analyzed in both original and enhanced images. Then, the color components of these two images are processed in order to keep the ratios of colors to be equal the average color ratios in images. The enhanced images and the images after color ratio correction (modification) represent in most cases high quality images that can be considered examples of restoration being close to the original images. Different measurements of the images are analyzed, including the new measurements that relate to relationships between the ratios of primary colors in the paintings. The analysis of different images calculated from the original image by using the enhancement methods and method of color ratio correction allow us to select the high-quality image and consider it as a predicted image for the painting.

  Now we describe the proposed method of color image processing in more detail. The color images are considered in the RGB color space, with the primary colors, red (R), green (G), and blue (B) in each pixel [30]. The discrete color image $f_{n,m}$ of size $N \times M$ pixels is map to the quaternion space, where the image can be presented as the quaternion image, or the 4-D image of size $N \times M$ pixels, that is defined as

$$f_{n,m} = a_{n,m} + r_{n,m}i + g_{n,m}j + b_{n,m}k. \quad (1)$$

Here, the imaginary units $i, j$, and $k$ satisfy the following multiplication rules:

$$ij = -ji = k, jk = -kj = i, ki = -ik = -j, \quad i^2 = j^2 = k^2 = ijk = -1.$$

The real part $a_{n,m}$ of the quaternion image usually is considered zero or the gray-scale component being the average $a_{n,m} = (r_{n,m} + g_{n,m} + b_{n,m})/3$. We consider the real part to be the brightness of the image that is calculated by

$$a_{n,m} = 0.3r_{n,m} + 0.59g_{n,m} + 0.11b_{n,m} \tag{2}$$

in each pixel $(n, m)$. When using quaternion operations, each color triple is treated as a whole unit, and in the traditional approach each color component is processed separately.

When enhancing color images in the frequency domain, the different concepts of the 2-D quaternion discrete Fourier transform (QDFT) can be used, since the multiplication in quaternion arithmetic is not commutative. Our experimental results show that the left, right, and both side 2-D QDFTs almost do not differ when using them in image enhancement [39]-[41]. Therefore, we consider the concept of the right-side 2-D QDFT. This transform of the quaternion image $f_{n,m}$ is calculated by

$$F_{p,s} = \sum_{n=0}^{N-1} \left( \sum_{m=0}^{M-1} f_{n,m} W_k^{ms} \right) W_j^{np}, \quad p = 0:(N-1), s = 0:(M-1). \tag{3}$$

The quaternion exponential functions are defined as

$$W_k^{ms} = \cos\left(\frac{2\pi ms}{M}\right) - k \sin\left(\frac{2\pi ms}{M}\right), \quad W_j^{np} = \cos\left(\frac{2\pi np}{N}\right) - j \sin\left(\frac{2\pi np}{N}\right).$$

Instead of the pure unit quaternions $j$ and $k$ in these basic exponential functions, other pure quaternions can also be used. The inverse 2-D right-side QDFT is calculated by

$$f_{n,m} = \frac{1}{NM} \sum_{p=0}^{N-1} \left( \sum_{s=0}^{M-1} F_{p,s} W_k^{-ms} \right) W_j^{-np}, \tag{4}$$

where $n = 0:(N-1)$ and $m = 0:(M-1)$. Different fast 2-D QDFTs were developed that are based on the fast 1-D QDFT, and the fast tensor 2-D QDFT [31]-[33]. Therefore, the 2-D QDFT can be effectively used for quaternion and color image processing in the frequency domain.

One of the effective methods of enhancing a quality of the color image is the method of alpha-rooting with the 2-D QDFT, when only magnitudes of the 2-D QDFT coefficients are changed by the real exponential function [34]-[36]. Given value of $\alpha$ from the interval (0,1), the $\alpha$-rooting is calculated by using the following steps:

1. The coefficients $F_{p,s}$ of the 2-D QDFT of the original color image are calculated.
2. The modules of these coefficients are modified by factors $C(p, s) = |F_{p,s}|^{\alpha-1}$, so that

$$|F_{p,s}| \to C(p,s)|F_{p,s}| = |F_{p,s}|^{\alpha}. \tag{5}$$

3. The new 2-D QDFT is calculated by $G_{p,s} = C(p,s)F_{p,s}$.
4. The inverse 2-D QDFT $g_{n,m}$ is calculated on the new transform.

The selection of best values of $\alpha$ for enhancing color images is based on using the known color image enhancement EMEC functions [20]-[23],[35]. We consider this measure of quality color images in the RGB model, wherein the color image $f_{n,m}$ is the triplet of red, green, and blue color components, $f = \{f_R, f_G, f_B\}$. The color image enhancement measure *EMEC* is calculated as follows:

$$EMEC(f) = \frac{1}{k_1 k_2} \sum_{k=1}^{k_1} \sum_{l=1}^{k_2} 20\log_{10}\left[\frac{\max_{k,l}(f_R, f_G, f_B)}{\min_{k,l}(f_R, f_G, f_B)}\right]. \tag{6}$$

Here, the discrete image is divided by $k_1 k_2$ blocks, where $k_1 = \lfloor N/L \rfloor$ and $k_2 = \lfloor M/L \rfloor$, the operation $\lfloor . \rfloor$ denotes the rounding floor function, and $L \times L$ is the size of blocks on which the mage is divided. In all examples given in this work, the block size is 7×7. The maximum value of the image in the $(k, l)$-th block is calculated by $\max(r_{n,m}, g_{n,m}, b_{n,m})$, i.e., component-wise. The minimum value in this block $\min(r_{n,m}, g_{n,m}, b_{n,m})$ is calculated similarly.

For a full quaternion image $f = \{f_A, f_R, f_G, f_B\}$, the enhancement measure *EMEQ* is calculated as follows [30,41]:

$$EMEQ(f) = \frac{1}{k_1 k_2} \sum_{k=1}^{k_1} \sum_{l=1}^{k_2} 20\log_{10}\left[\frac{\max_{k,l}(f_A, f_R, f_G, f_B)}{\min_{k,l}(f_A, f_R, f_G, f_B)}\right]. \quad (7)$$

with maximum and minimum values of the image in the $(k,l)$-th block that are calculated by $\max(a_{n,m}, r_{n,m}, g_{n,m}, b_{n,m})$ and $\min(a_{n,m}, r_{n,m}, g_{n,m}, b_{n,m})$, i.e., these operations include the real part of the quaternion image.

According to the definition of the EMEQ, this measure calculates the average of the range of color in the logarithmic scale. The measure EMEQ is calculated for the original color image $f_{n,m}$ and enhanced image $g_{n,m}$. Usually, the visually enhanced image has high value of the measure. In alpha-rooting, the enhanced image is parameterized by $\alpha$, the same is the enhancement measure. Therefore, the best parameter for color enhancement is considered the value of $\alpha$ with maximum of $EMEQ(g)$. Our experimental results show that the measures EMEQ and EMEC are effective in selecting the best parameters to receive images with high quality [38,41]. Other measures for estimating image quality can also be used, including color image contrast and quality measures [27]-[29].

To estimate the relationship between the primary colors in the RGB model, we introduce the following three measures:

1. Measurement of red-to-blue colors

$$M_1(f) = M_{r/b}(f) = \frac{1}{NM} \sum_{n=0}^{N-1} \sum_{m=0}^{M-1} \frac{\log_{10}(r_{n,m}) - \log_{10}(\text{Mean}[r])}{\log_{10}(b_{n,m}) - \log_{10}(\text{Mean}[b]) + \varepsilon}. \quad (8)$$

2. Measurement of green-to-blue colors

$$M_2(f) = M_{g/b}(f) = \frac{1}{NM} \sum_{n=0}^{N-1} \sum_{m=0}^{M-1} \frac{\log_{10}(g_{n,m}) - \log_{10}(\text{Mean}[g])}{\log_{10}(b_{n,m}) - \log_{10}(\text{Mean}[b]) + \varepsilon}. \quad (9)$$

3. Measurement of blue-to-red colors

$$M_3(f) = M_{b/r}(f) = \frac{1}{NM} \sum_{n=0}^{N-1} \sum_{m=0}^{M-1} \frac{\log_{10}(b_{n,m}) - \log_{10}(\text{Mean}[b])}{\log_{10}(r_{n,m}) - \log_{10}(\text{Mean}[r]) + \varepsilon}, \quad (10)$$

where $\varepsilon$ is a small number, such as 0.001. Here $\text{Mean}[r]$, $\text{Mean}[g]$, and $\text{Mean}[b]$ are the mean values of the red, green, and blue color components of the image, respectively. As the mean average of these measurements, we also consider the measure

$$M(f) = \sqrt[3]{M_{r/b}(f) M_{g/b}(f) M_{b/r}(f)}. \quad (11)$$

## 3. COLORS AND SYMBOLS

In this section, we briefly consider the symbolic interpretation of main colors that was used in Byzantine, Gothic and Renaissance paintings. The following colors were used in Renaissance art [8]-[15].

1. **Gold** - used as a background color or in a halo symbolizes purity, royalty and glory of life after death. Gold is associated with wealth, royalty and heavenly rewards and riches.
2. **Blue** - symbolizes purity; The Virgin Mary; Virgin and Child; The Immaculate Conception.
3. **Purple** - symbolizes Christ in Majesty, In Byzantine Style painting Important Holy figures wear purple robes, outlined in red.
4. **Red** - a symbol of greediness and lust. Denotes sin; sins of mankind, original sin; temptation, Judas, Harrowing of Hell, The Fiery Furnace, Slaughter of the Innocents, Apocalypse. Red also denotes power and authority.

5. **Green** - symbolizes the Resurrection, the Ascension, and Baptism. Green also symbolizes peace, spring, spiritual renewal, rebirth and new life
6. **White** - is a sign of innocence; Birth, Youth, Betrothal and Marriage; The Virgin Mary; Virgin and Child; The Immaculate Conception, The Holy Family and angels.
7. **Grey/Black/Dark Brown** - symbolic of the Entombment, Crosses, Crucifixes as well as darkness, misdeeds, death, witchery.
8. **Yellow** - symbolizes a remembrance of the spiritual world; miracles, harmony, sustenance of the soul. Yellow also symbolizes spring, spiritual renewal, rebirth and new life.
9. **Pink** - symbolizes eternal innocence; The Virgin Mary; Virgin and Child; The Immaculate Conception
10. **Orange** - a symbol of materialism and desire for worldly goods in favour of spiritual health. Denotes indulgence; carnal desires, original sin.

In the Renaissance, the colors were used and selected based on the several factors that include the following:
1. Color costs: for example deep blue color ingredients were so expensive that they were saved for only special parts of a picture (for the clothing of the biblical Mary, or a woman wearing clothes), because they were made from minerals, such as gemstone lapis lazuli into a fine powder and mixing it with other ingredients
2. Aesthetic or Technical Purposes:
    o to create the appearance of a three-dimensional space, which means making a scene look as if you could almost walk into it.
    o color was often chosen simply it also served as a symbol in selected cases
    o context determines the meaning of color
3. Local Culture and Geographic Area.

Below is Table that presents the meaning of the renaissance color symbolism.

| Color and its properties | Renaissance | Medieval |
|---|---|---|
| Reds<br>Wavelength interval of 700-635 nm<br>Frequency interval of 430-480 THz<br>RGB (255,0,0) | • Is associated with High social status, royalty, gentlemen, men of justice (Scotland, the Holy Roman Empire, England's Court of Common Pleas, occasionally by peers in English Parliament or, man with access to international trading centres) [1,3]<br>• Is associated with high government posts (Venice and Florence), royal magistrates, and king's chancellor (France);<br>• Is associated with Power and prestige [2].<br>• Is associated with is authority, Pentecostal fire, the blood of Christ, martyrdom, crucifixion, Christian charity.<br>• Is symbolized the satanic and color of hellfire [2]<br>• Is associated with medicine. (At the universities of Padua and Bologna) [1] | • A lover wears vermilion, like blood' (later Middle Ages) [4]<br>• A sign of otherworldly power in European legends and folktales.<br>• Protection: red thread to ward off witches, red coral necklaces to guard against illness [2].<br>• The Virgin Mary's robes [2]<br>• Identified with kingly virtues of<br>• Valour and success in war [2]<br>• Fire [2] |

| Color and its properties | Renaissance | Medieval |
|---|---|---|
| 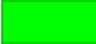 Green<br>Wavelength interval of 560-520 nm<br>Frequency interval of 540-580 THz<br>RGB (0,255,0) | Symbolizes Love and Joy [2]<br>Associates with Youth [1]<br>Associates within the secular sphere, chastity [3] | Nothing noted. |
| 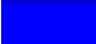 Blue<br>Wavelength interval of 490-450 nm<br>Frequency interval of 610-670 THz<br>RGB (0,255,0) | • Light blue represented a young marriageable woman [1]<br>• It was the traditional color of servitude (England). Servants or members of a City company were to wear bright blue or gray Renaissance clothing [7]<br>• Deep blue associates with chastity in the sacred sphere [3]<br>• ". . . turquoise was a sure sign of jealousy . . ." [2] | Middle Ages:<br>• Blue replaced royal purple in the mantle of the Virgin Mary and robes and heraldry (especially in France) [15]<br>• A lover wears blue for fidelity (late) [4]<br>• Blue was associated with darkness, evil. Later blue was associated with light [1] |
| 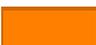 Orange<br>Wavelength interval of 635-590 nm<br>Frequency interval of 480-510 THz<br>RGB (0,255,0) | Is associated with peasants and middle ranked persons imitated upper class reds by dyeing their Renaissance clothes with cheaper orange-red and russet dyes [2]. | Illness |
| 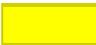 Yellow<br>Wavelength interval of 590-560 nm<br>Frequency interval of 510-540 THz<br>RGB (255,255,0) | Is associated with prostitution (In almost all Italian cities, was required to wear yellow [1]<br>Is associated with Jews (In Venice were required to sew a yellow circle onto clothing [1]<br>is very useful in attracting attention | Color expressing the balance between the red of justice and the white of compassion [5] |
| 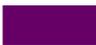 Purple<br>Wavelength interval of 450-400 nm<br>Frequency interval of 670-750 THz<br>RGB (255,255,0) | Is associated with the Medici family in Florence [1] (Imperial purple disappeared in 1453) [2]. | |
| Gray | Is associated with<br>a) The modest and religious dress [9]<br>b) The poverty [2]<br>c) The lower classes) (for example In England, servants or members of a City company were to wear bright blue or gray [9]. | |
| 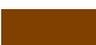 Brown<br>RGB (255,255,0) | Is associated with<br>a) Modest and religious dress [2]<br>b) The color of poverty [2]<br>c) Dull browns were worn by lower classes (In England) [2] | |

| Color and its properties | Renaissance | Medieval |
|---|---|---|
| 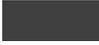 Black | is associated with<br>a) Seriousness [11]<br>b) Mourning [1]<br>c) Nobility and wealthy, representing refinement and distinction [2] | |

## 4. THE COLOR GOLDEN RATIO

The analysis of many images of paintings of well-known artist's shows that many artists had their unique ratio in colors. Such ratios include for instance the well-known golden ratio (GR) [16]-18], and effectively used in signal and image processing [24]-[26]. The GR is a rule of the proportionality of the whole and the parts that compose the whole. For intervals of lengths $a$ and $b$, when $b > a > 0$, the GR is described by the following mean-proportional relation

$$\frac{b+a}{b} = \frac{b}{a} = \Psi, \tag{12}$$

The GR is one of solutions of the equation $\Psi^2 - \Psi - 1 = 0$ and equal to

$$\Psi = (1+\sqrt{5})/2 = 1.618033988\ldots.$$

We call three numbers $x, y, z$ in gold proportion, if one of the following relations is valid:

$$\begin{aligned}
(x,y,z) &= a \times (\Psi, \sqrt{\Psi}, 1), & (x,z,y) &= a \times (\Psi, \sqrt{\Psi}, 1), & (y,x,z) &= a \times (\Psi, \sqrt{\Psi}, 1), \\
(y,z,x) &= a \times (\Psi, \sqrt{\Psi}, 1), & (z,y,x) &= a \times (\Psi, \sqrt{\Psi}, 1), & (z,x,y) &= a \times (\Psi, \sqrt{\Psi}, 1),
\end{aligned} \tag{13}$$

where $a$ is a positive number, and $\Phi = \sqrt{\Psi} = 1.2720$. If the numbers $x, y, z$ preset the color components $r, b, g$, then the color $(r, g, b)$ is called the golden color. Together with the number $\Psi$, we can describe similar definitions of colors in other well-known ratios that include the Aesthetic 1.322, Silver Mean $1 + \sqrt{2} = 2.4142$, Copper Mean 2, Bronze Mean $(3 + \sqrt{13})/2 = 3.3027$, Nickel Mean $1 + \sqrt{13} = 5.6055$ and others [19].

It is assumed that each recognized artist has its color ratio, which we denote by $\Phi$, not necessary being the number $\sqrt{\Psi} = 1.2720$ of GR or other mentioned above ratios. This ratio can be estimated by the average of colors when analyzing a few dozens of paintings of the artist. For instance, our research show that for Leonardo Da Vinci $\Phi = 1.46$, for Pablo Picasso $\Phi = 1.49$, for Vincient van Gogh $\Phi = 1.38$, for Rafaello Sanzio $\Phi = 1.61$, and Rembrandt van Rijn $\Phi = 1.65$. Then, the question arises how the images of paintings look after correction of color components at all pixels, in such way that colors will be in $\Psi$-proportion that is described by equations in (10).

The method of color correction for the given $\Psi$-proportion is simple. First, we consider the permutation of colors $(r, g, b) \to (x, y, z)$ in descending order for each pixel $(n, m)$. Then, the colors are processed as $(x, y, z) \to (x_n, y_n, z_n)$, where the new image the colors $x', y'$, and $z'$ satisfy the condition $x'/y' = y'/z'$. For that, the following calculations are performed $x_n = x$, $y_n = y$, and the value of $z_n$ is calculated by $z_n = y/\psi$, where $\psi = x_n/y_n$. Therefore, $x_n/y_n = y_n/z_n = \psi$. This method of color correction is called *the color modeling via color ratio* (CMCR).

To illustrate the proposed method of processing of color images, Fig. 2 shows the color image of size 532×729 pixels of one of Leonardo Da Vinci's paintings and nine images of α-rooting by the 2-D QDFT for α=0.80:0.96 with step 0.02.

The mean of the ratios $\psi$ of the image $f_{n,m}$ is calculated by

$$CR = CR(f) = \text{Mean}(\psi) = \frac{1}{NM} \sum_{n=0}^{N-1} \sum_{m=0}^{M-1} \psi_{n,m} \qquad (14)$$

where $\psi_{n,m}$ is the color ratio at pixel $(n, m)$.

As an example, Figure 2 shows the original image of the "Tobit and Anna" of Rembrandt's painting and the alpha-rooting by the 2-D QDFT, when the values of alpha α=0.80, 0.82, 0.84, 0.86, 0.88, 0.90, 0.92, 0.94, and 0.96. One can note the quality of the images of 0.92-rooting for which the measurements CR and M1 have high values.

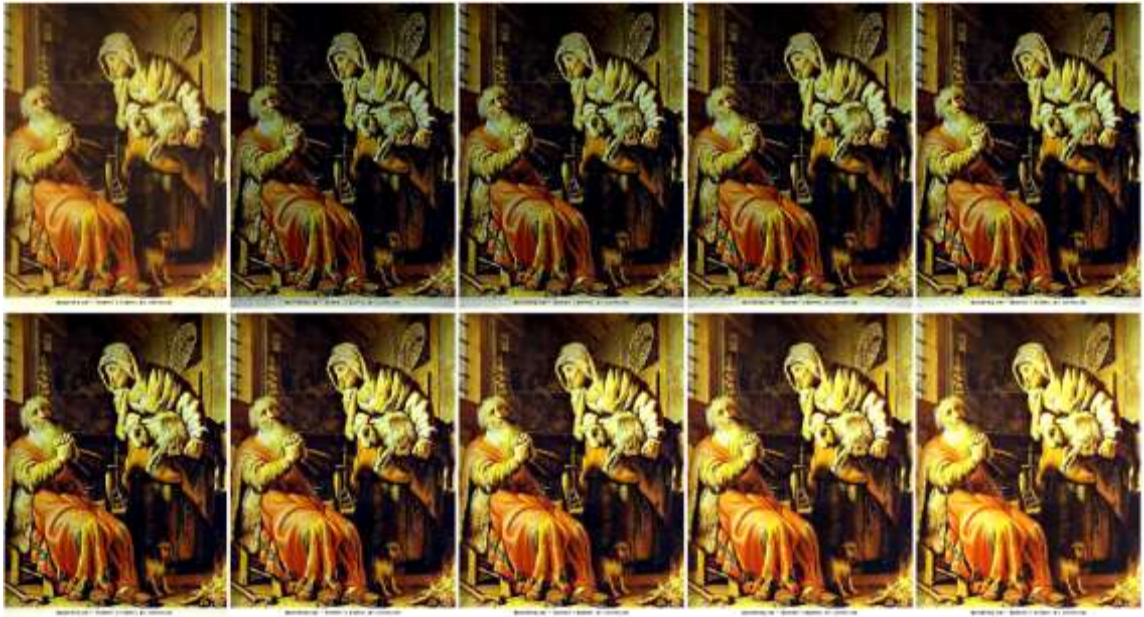

**Fig. 2** (from the left to right, top to bottom) The original color image [rembrandt1.jpg] of size 802×586 pixels (from http://www.abcgallery.com/R/rembrandt/rembrandt.html) and α-rooting by the 2-D QDFT for α=0.80, 0.82, 0.84, 0.86, 0.88, 0.90, 0.92, 0.94, and 0.96.

Data of the color ratio and the measurements for the images of Fig. 2 are given in Table 1. For each of these images, the measure EMEC of the image is calculated also after modifying the color ratio to the mean ratio. This measure is denoted in the table by EMEC2. The images of 0.88- and 0.92-rootings with maximum measures CR and M has very good quality and can be considered as images of predicted colors in this painting.

| α | 0.80 | 0.82 | 0.84 | 0.86 | 0.88 | 0.90 | 0.92 | 0.94 | 0.96 | 0.98 | 1.00 |
|---|------|------|------|------|------|------|------|------|------|------|------|
| CR | 1.83 | 1.89 | 1.94 | 1.98 | **2.02** | 2.01 | **1.96** | 1.86 | 1.75 | 1.65 | 1.58 |
| M1 | 1.14 | 1.14 | 1.36 | 1.15 | 1.30 | 1.31 | 1.87 | 1.18 | 1.04 | 1.04 | 1.08 |
| M2 | 1.07 | 1.10 | 1.30 | 1.14 | 1.24 | 1.27 | 1.64 | 1.16 | 1.06 | 1.05 | 1.05 |
| M3 | 1.62 | 2.61 | 2.97 | 5.59 | 3.17 | 3.52 | 3.39 | 3.37 | 3.54 | 5.05 | 3.68 |
| M | 1.25 | 1.49 | 1.74 | 1.94 | 1.73 | 1.80 | **2.19** | 1.66 | 1.57 | 1.77 | 1.61 |
| EMEC | 2.66 | 3.14 | 3.77 | 4.69 | 6.36 | 8.18 | 11.60 | 15.72 | 19.57 | 23.67 | 27.56 |
| EMEC2 | 5.47 | 6.29 | 7.33 | 8.83 | 11.12 | 14.76 | 19.36 | 24.87 | 29.57 | 33.43 | 35.74 |

**Table 1.** Data for the images in Fig. 2.

For the image of the Rembrandt's painting "Tobit and Anna," Fig. 3 shows the graph of calculation of the color ratio of the color images after alpha-rooting by the 2-D QDFT. The color ratio as the function of alpha, CR(α), is shown in the interval [0,7,1] with step 0.01. The maximum of the color ratio is at point $\alpha_0 = 0.89$ with the values 2.0205.

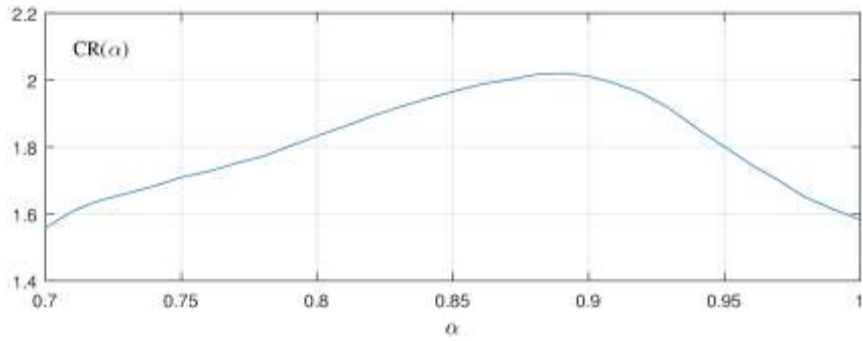

**Fig. 3** The graph of the color ratio in the interval [0,7,1] with step 0.01.

Figure 4 shows the original image in part (a), the 0.89-rooting by the 2-D QDFT in part (b). The average color ratio of the image in (b) equals CR=2.0205, and after reducing the color ratio to this image to value 2.0205 is shown in part (c). Both images have a good quality and color pallete. It should be noted for comparison that the original image has color ratio equal CR=2.4387 and, after reducing the color ratio to this value, this image is shown in part (e).

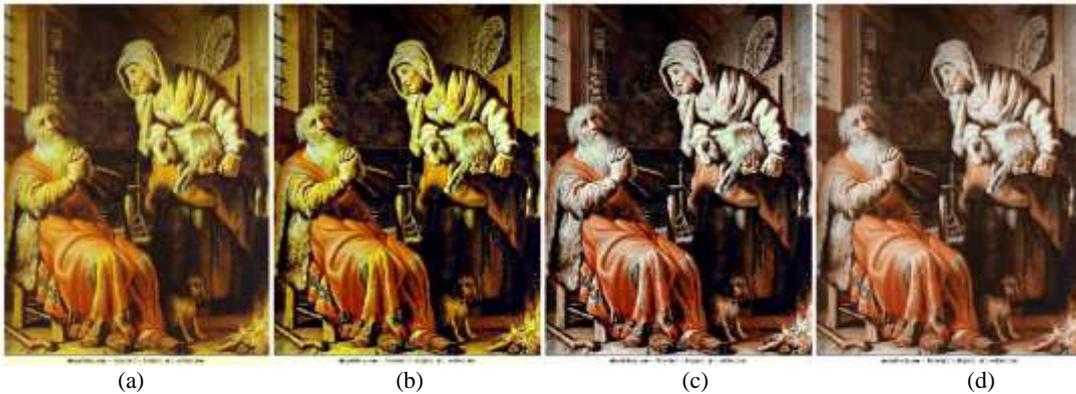

(a) (b) (c) (d)

**Fig. 4** (a) The original color image [rembrandt1.jpg] (from http://www.abcgallery.com/R/rembrandt/rembrandt.html), (b) 0.89-rooting by the 2-D QDFT, (c) the 0.89-rooting when changing the color ratio to 2.0205, and (d) the original image with the color ratio 2.4387.

Figure 5 shows the original image of the "Madonna Litta" of Leonardo Da Vinci's painting and the alpha-rooting by the 2-D QDFT, for nine values of alpha α=0.80, 0.82, 0.84, ..., 0.92, 0.94, and 0.96. One can notice the increase in quality of images when changing value of alpha after the point 0.90.

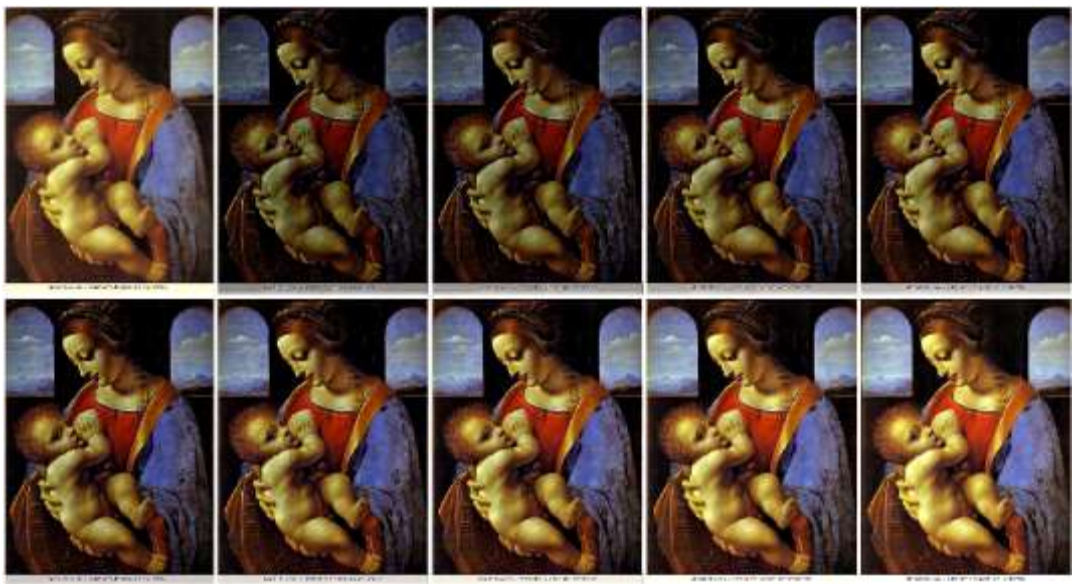

**Fig. 5** (from the left to right, top to bottom) The original image [leonardo13.jpg] of size 729×532 pixels (from http://www.abcgallery.com/L/leonardo/leonardo.html) and α-rooting by the 2-D QDFT for α=0.80, 0.82, 0.84, 0.86, 0.88, 0.90, 0.92, 94, and 0.96.

Data of the color ratio and the measurements for the images of Fig. 5 are given in Table 2.

| α | 0.80 | 0.82 | 0.84 | 0.86 | 0.88 | 0.90 | 0.92 | 0.94 | 0.96 | 0.98 | 1.00 |
|---|---|---|---|---|---|---|---|---|---|---|---|
| CR | 1.96 | 2.02 | 2.05 | 2.04 | 2.02 | 1.97 | 1.91 | 1.84 | 1.76 | 1.64 | 1.56 |
| M1 | 1.48 | 1.84 | 1.07 | 1.07 | 1.16 | 1.20 | 1.19 | 1.10 | 1.17 | 1.06 | 1.53 |
| M2 | 1.39 | 1.65 | 1.07 | 1.08 | 1.15 | 1.18 | 1.18 | 1.10 | 1.14 | 1.04 | 1.36 |
| M3 | 2.09 | 5.09 | 4.31 | 3.57 | 4.40 | 5.25 | 4.74 | 5.57 | 7.66 | 6.54 | 5.24 |
| M | 1.63 | 2.49 | 1.70 | 1.60 | 1.80 | 1.96 | 1.88 | 1.89 | **2.17** | 1.93 | 2.22 |
| EMEC | 7.28 | 7.68 | 8.28 | 9.32 | 10.41 | 12.17 | 14.62 | 17.45 | 21.08 | 23.87 | 25.72 |
| EMEC2 | 7.27 | 8.35 | 9.26 | 10.17 | 11.45 | 13.99 | 17.08 | 20.62 | 23.74 | 27.17 | 29.98 |

Table 2: Data for the images in Fig. 5.

Figure 6 shows the color image of the "Virgin and Child with St. Anne" of Leonardo Da Vinci's painting and the alpha-rooting by the 2-D QDFT, for nine values of alpha α=0.82, 0.84, ..., 0.94, 0.96, and 0.98. We can note that the images of 0.90- and 0.92-rooting show high quality images; all colors and details in the images can be seen in a wonderful color pallete. The measures M and CR have the high values for the 0.90-rooting.

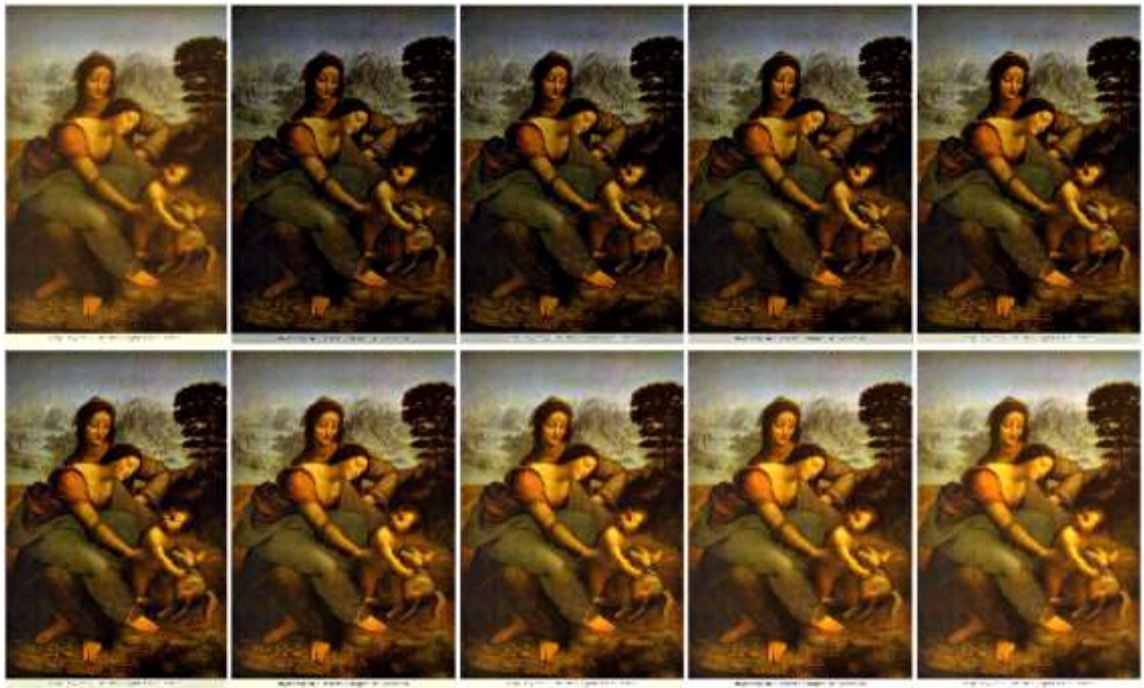

**Fig. 6** (from the left to right, top to bottom) The original color image [leonardo3.jpg] of size 729×532 pixels (from http://www.abcgallery.com/L/leonardo/leonardo.html) and α-rooting by the 2-D QDFT for α= 0.82, 0.84, 0.86, 0.88, 0.90, 0.92, 94, 0.96, and 0.98.

Data of the color ratio and the measurements for the images of Fig. 6 are given in Table 3.

| α | 0.80 | 0.82 | 0.84 | 0.86 | 0.88 | 0.90 | 0.92 | 0.94 | 0.96 | 0.98 | 1.00 |
|---|---|---|---|---|---|---|---|---|---|---|---|
| CR | 1.76 | 1.77 | 1.77 | 1.76 | 1.77 | **1.78** | 1.75 | 1.65 | 1.55 | 1.49 | 1.46 |

| M1 | 1.19 | 1.37 | 1.07 | 1.55 | 1.04 | 2.97 | 2.29 | 3.64 | 1.92 | 1.08 | 1.60 |
| M2 | 1.04 | 1.15 | 0.99 | 1.24 | 0.97 | 1.99 | 1.62 | 2.33 | 1.41 | 0.97 | 1.24 |
| M3 | 1.90 | 2.46 | 2.76 | 2.98 | 3.40 | 19.33 | 5.26 | 4.26 | 4.02 | 4.51 | 4.39 |
| M | 1.33 | 1.57 | 1.43 | 1.79 | 1.51 | **4.85** | 2.69 | 3.31 | 2.22 | 1.67 | 2.06 |
| EMEC | 5.26 | 5.89 | 6.61 | 7.75 | 8.95 | 10.47 | 12.62 | 15.52 | 18.48 | 22.23 | 26.87 |
| EMEC2 | 8.07 | 9.30 | 11.32 | 13.70 | 16.29 | 19.35 | 22.93 | 26.16 | 28.06 | 26.74 | 24.68 |

**Table 3:** Data for the images in Fig. 6.

Figure 7 shows the color image of the "Allegory (The Knight's Dream)" of Raphael's painting and the alpha-rooting by the 2-D QDFT, for nine values of alpha α=0.80, 0.82, 0.84, ..., 0.92, 0.94, and 0.96. One can notice that the color pallete changes slowly with increase of values of alpha.

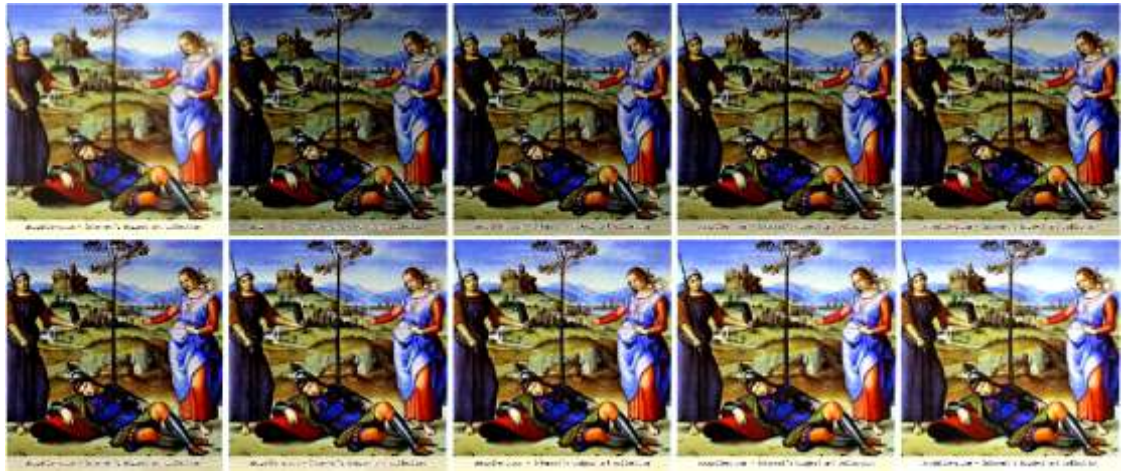

**Fig. 7** (from the left to right, top to bottom) The original color image [srafael11.jpg] of size 398×373 pixels (from http://www.abcgallery.com/r/raphael/raphael.html) and α-rooting by the 2-D QDFT for α=0.80, 0.82, 0.84, 0.86, 0.88, 0.90, 0.92, 0.94, and 0.96.

Data of the color ratio and the measurements for the images of Fig. 7 are given in Table 4.

| α | 0.80 | 0.82 | 0.84 | 0.86 | 0.88 | 0.90 | 0.92 | 0.94 | 0.96 | 1 |
|---|---|---|---|---|---|---|---|---|---|---|
| CR | 2.26 | 2.34 | 2.39 | 2.49 | 2.53 | **2.53** | 2.43 | 2.23 | 1.92 | 1.50 |
| M1 | 1.48 | 1.58 | 4.73 | 1.69 | 1.79 | 1.93 | 1.85 | 1.89 | 1.91 | 2.35 |
| M2 | 1.47 | 1.57 | 4,77 | 1.69 | 1.81 | 1.98 | 1.91 | 1.99 | 2.02 | 2.58 |
| M3 | 2.98 | 2.11 | 5.33 | 2.49 | 3.04 | 7.17 | 3.20 | 5.60 | 6.53 | 4.62 |
| M | 1.87 | 1.73 | 4.94 | 1.92 | 2.14 | **3.02** | 2.24 | 2.76 | 2.93 | 3.03 |
| EMEC | 7.25 | 7.84 | 8.13 | 8.56 | 9.10 | 9.78 | 11.19 | 12.69 | 15.52 | 23.08 |
| EMEC2 | 8.09 | 9.03 | 10.30 | 11.13 | 11.97 | 13.69 | 14.69 | 17.15 | 22.16 | 35.42 |

Table 4: Data for the images in Fig. 7.

Figure 8 shows the color image of the "Canigiani Holy Family" of Raphael's painting and the alpha-rooting by the 2-D QDFT, for five values of alpha α=0.90, 0.92, 0.94, 0.96, and 0.98.

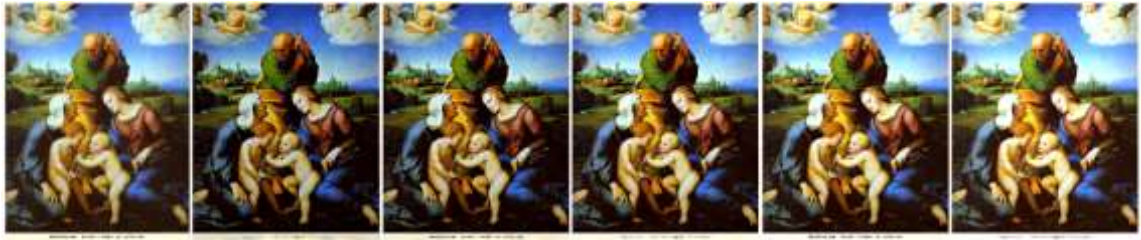

**Fig. 8** (from the left to right, top to bottom) The original color image [srafael25.jpg] of size 771×599 pixels (from http://www.abcgallery.com/r/raphael/raphael.html) and α-rooting by the 2-D QDFT for α=0.90, 0.92, 0.94, 0.96, and 0.98.

Data of the color ratio and the measurements for these images of the alpha-tooting are given in Table 5.

| α | 0.80 | 0.82 | 0.84 | 0.86 | 0.88 | 0.90 | 0.92 | 0.94 | 0.96 | 0.98 | 1.00 |
|---|---|---|---|---|---|---|---|---|---|---|---|
| CR | 1.91 | 1.86 | 1.82 | 1.83 | 1.90 | 2.03 | 2.23 | **2.27** | 1.99 | 1.64 | 1.50 |
| M1 | 1.39 | 6.55 | 3.29 | 1.53 | 1.54 | 1.70 | 1.51 | 1.62 | 1.41 | 1.78 | 1.70 |
| M2 | 1.28 | 5.18 | 2.56 | 1.26 | 1.27 | 1.40 | 1.29 | 1.40 | 1.27 | 1.54 | 1.46 |
| M3 | 1.72 | 2.12 | 2.25 | 2.82 | 4.72 | 3.09 | 3.28 | 3.66 | 4.38 | 4.16 | 3.48 |
| M | 1.41 | 4.16 | 2.67 | 1.75 | 2.10 | 1.95 | 1.86 | **2.02** | 1.98 | **2.25** | 2.05 |
| EMEC | 8.12 | 8.20 | 8.52 | 8.75 | 9.31 | 10.07 | 11.49 | 13.79 | 17.24 | 21.09 | 23.97 |
| EMEC2 | 8.45 | 9.60 | 10.56 | 11.55 | 12.41 | 13.32 | 14.71 | 17.21 | 23.39 | 27.74 | 27.74 |

Table 5: Data of processing the images of the painting "Canigiani Holy Family."

Figure 9 shows the color image of the "Madonna and Child" of Raphael's painting and the alpha-rooting by the 2-D QDFT, for five values of alpha α=0.82, 0.86, 0.90, 0.94, and 0.98.

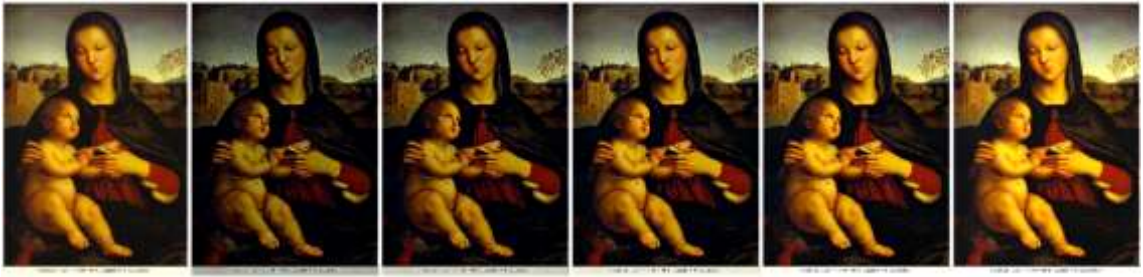

**Fig. 9** (from the left to right, top to bottom) The original color image [srafael14.jpg] of size 735×500 pixels (from http://www.abcgallery.com/r/raphael/raphael.html) and α-rooting by the 2-D QDFT for α=0.82, 0.86, 0.90, 0.94, and 0.98.

Data of the color ratio and the measurements for the images of Fig. 9 are given in Table 6.

| α | 0.80 | 0.82 | 0.84 | 0.86 | 0.88 | 0.90 | 0.92 | 0.94 | 0.96 | 0.98 | 1.00 |
|---|---|---|---|---|---|---|---|---|---|---|---|
| CR | 1.96 | 2.09 | 2.20 | 2.24 | 2.28 | **2.30** | 2.20 | 2.03 | 1.81 | 1.69 | 1.64 |
| M1 | 1.73 | 13.19 | 2.13 | 1.46 | 1.82 | 1.98 | 1.77 | 1.65 | 1.51 | 1.45 | 1.51 |
| M2 | 1.44 | 8.85 | 1.69 | 1.30 | 1.52 | 1.58 | 1.45 | 1.38 | 1.27 | 1.21 | 1.22 |
| M3 | 1.49 | 2.39 | 2.30 | 2.90 | 3,29 | 3.86 | 7.14 | 4.08 | 4.84 | 4.90 | 4.86 |
| M | 1.55 | 6.53 | 2.02 | 1.77 | 2.09 | 2.30 | **2.64** | 2.10 | 2.10 | 2.05 | 2.08 |
| EMEC | 3.29 | 3.49 | 3.58 | 3.87 | 4.55 | 5.49 | 7.41 | 10.79 | 14.57 | 17.48 | 19.89 |
| EMEC2 | 6.35 | 6.85 | 7.49 | 8.00 | 9.16 | 10.73 | 13.61 | 19.46 | 24.06 | 27.31 | 28.91 |

Table 6: Data of processing the images of the painting "Madonna and Child."

Figure 10 shows the color image of the 'Colonna Madonna' of Raphael's painting and the alpha-rooting by the 2-D QDFT, for nine values of α=0.82, 0.86, 0.90, 0.94, and 0.98.

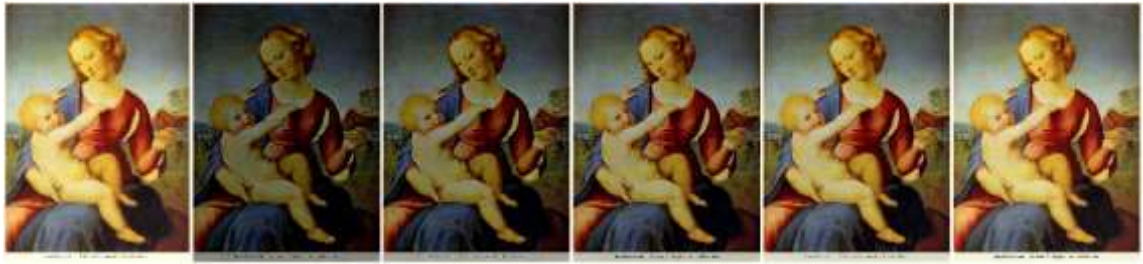

**Fig. 10** (from the left to right, top to bottom) The original color image [srafael91.jpg] of size 748×537 pixels (from http://www.abcgallery.com/r/raphael/raphael.html) and α-rooting by the 2-D QDFT α=0.82, 0.86, 0.90, 0.94, and 0.98.

Data of the color ratio and the measurements for the images of Fig. 10 are given in Table 7.

| α | 0.80 | 0.82 | 0.84 | 0.86 | 0.88 | 0.90 | 0.92 | 0.94 | 0.96 | 0.98 | 1 |
|---|---|---|---|---|---|---|---|---|---|---|---|
| CR | 1.90 | 1.96 | 2.07 | 2.18 | 2.30 | 2.40 | **2.44** | 2.32 | 2.01 | 1.68 | 1.49 |
| M | 1.58 | 2.33 | 2.13 | 1.98 | 2.26 | 2.46 | 2.23 | 2.45 | 2.53 | 2.73 | 2.97 |
| M1 | 1.48 | 1.96 | 1.66 | 1.50 | 1.61 | 1.64 | 1.48 | 1.53 | 1.50 | 1.55 | 1.63 |
| M2 | 1.60 | 4.61 | 1.77 | 5.79 | 5.23 | 2.75 | 5.54 | 4.27 | 5.10 | 8.12 | 7.89 |
| M | 1.55 | 2.76 | 1.84 | 2.58 | 2.67 | 2.23 | 2.64 | 2.52 | 2.69 | 3.25 | 3.37 |
| EMEC | 11.23 | 11.49 | 11.91 | 12.33 | 12.79 | 13.26 | 13.89 | 14.65 | 16.39 | 18.80 | 20.46 |
| EMEC2 | 10.91 | 12.26 | 12.52 | 12.76 | 12.82 | 13.23 | 13.54 | 14.83 | 16.86 | 19.21 | 20.98 |

Table 7: Data of processing the images of the painting "Colonna Madonna."

Figure 11 shows the color image of the "The Niccolini-Cowper Madonna" of Raphael's painting and the alpha-rooting by the 2-D QDFT, for nine values of α=0.82, 0.86, 0.90, 0.94, and 0.98.

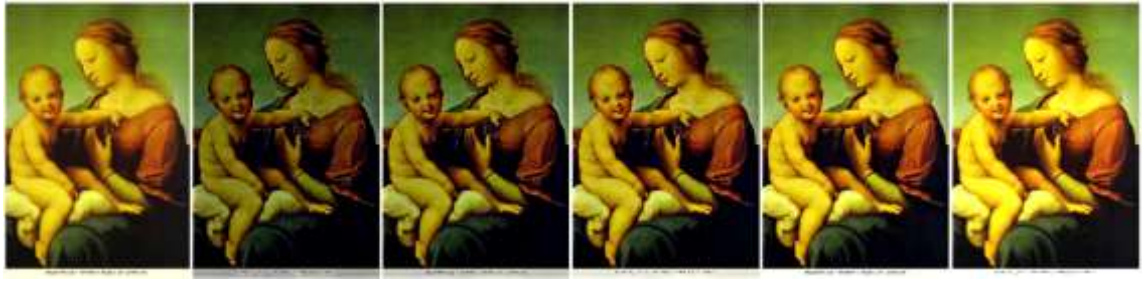

**Fig. 11** (from the left to right, top to bottom) The original color image [srafael94.jpg] of size 819×555 pixels (from http://www.abcgallery.com/r/raphael/raphael.html) and α-rooting by the 2-D QDFT α=0.82, 0.86, 0.90, 0.94, 0.98.

Data of the color ratio and the measurements for the images of Fig. 11 are given in Table 8.

| α | 0.80 | 0.82 | 0.84 | 0.86 | 0.88 | 0.90 | 0.92 | 0.94 | 0.96 | 0.98 | 1 |
|---|---|---|---|---|---|---|---|---|---|---|---|
| CR | 1.86 | 1.89 | 1.93 | 1.96 | 1.98 | **1.99** | 1.98 | 1.88 | 1.70 | `.52 | 1.45 |
| M1 | 1.07 | 1.16 | 12.05 | 4.19 | 2.62 | 2.61 | 5.49 | 1.68 | 2.13 | 1.83 | 2.20 |
| M2 | 1.04 | 1.13 | 14.75 | 5.40 | 3.28 | 3.32 | 6.42 | 1.72 | 1.97 | 1.56 | 1.59 |
| M3 | 1.57 | 1.58 | 1.67 | 2.00 | 3.44 | 12.44 | 4.40 | 3.37 | 4.74 | 3.75 | 4.04 |
| M | 1.20 | 1.28 | 6.67 | 3.56 | 3.09 | 4.76 | **5.37** | 2.14 | 2.71 | 2.20 | 2.42 |
| EMEC | 8.15 | 9.51 | 10.82 | 12.06 | 13.19 | 13.90 | 15.57 | 18.18 | 20.78 | 22.20 | 25.52 |
| EMEC2 | 9.99 | 11.33 | 11.57 | 12.12 | 13.03 | 13.97 | 15.40 | 17.56 | 20.17 | 21.12 | 21.03 |

Table 8: Data of processing the images of the painting "The Niccolini-Cowper Madonna."

Figure 12 shows the color image of the 'St. Catherine' of Raphael's painting and the alpha-rooting by the 2-D QDFT, for nine values of α=0.82, 0.86, 0.90, 0.94, and 0.98.

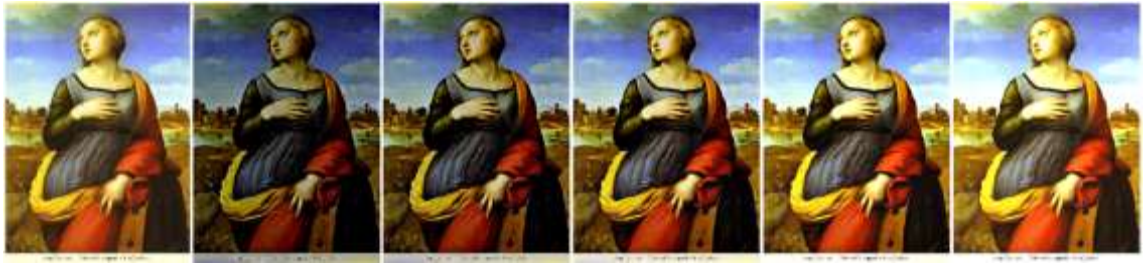

**Fig. 12** (from the left to right, top to bottom) The original color image [srafael27.jpg] of size 741×528 pixels (from http://www.abcgallery.com/r/raphael/raphael.html) and α-rooting by the 2-D QDFT α=0.82, 0.86, 0.90, 0.94, 0.98. Data of the color ratio and the measurements for the images of Fig. 12 are given in Table 9.

| α | 0.80 | 0.82 | 0.84 | 0.86 | 0.88 | 0.90 | 0.92 | 0.94 | 0.96 | 0.98 | 1.00 |
|---|------|------|------|------|------|------|------|------|------|------|------|
| CR | 2.15 | 2.21 | 2.26 | **2.29** | 2.28 | 2.21 | 2.05 | 1.96 | 1.68 | 1.57 | 1.51 |
| M1 | 1.36 | 1.46 | 1.20 | 1.47 | 1.28 | 3.27 | 1.15 | 1.65 | 1.29 | 1.14 | 1.06 |
| M2 | 1.34 | 1.42 | 1.27 | 1.39 | 1.21 | 2.85 | 1.12 | 1.56 | 1.27 | 1.16 | 1.09 |
| M3 | 7.65 | 2.40 | 6.32 | 3.88 | 4.73 | 3.74 | 4.52 | 4.96 | 5.32 | 6.41 | 7.27 |
| M | 2.41 | 1.71 | 2.19 | 1.99 | 1.95 | **3.27** | 1.80 | 2.34 | 2.06 | 2.03 | 2.03 |
| EMEC | 8.74 | 8.38 | 8.34 | 8.53 | 8.81 | 9.24 | 10.96 | 12.98 | 15.79 | 20.15 | 25.14 |
| EMEC2 | 10.29 | 11.59 | 13.15 | 14.53 | 15.50 | 16.87 | 19.40 | 22.92 | 26.00 | 26.13 | 24.09 |

Table 9: Data of processing the images of the painting "St. Catherine."

The images of 0.86- and 0.90-rootings have high values of the main measures CR and M in this analysis and visually also one can note that these images are of high quality and probably are the best candidate to select for color prediction for this painting.

## 6. Summary Results

The images are taken from Olga's Gallery by address: "http://www.abcgallery.com/." 105 images of Rafael, Leonardo Da Vinci, and Rembrandt's paintings were selected and processed. The data of color ratio and discussed above measurements were calculated for each image as well as these measurements were averaged. These data are shown in Tables 10-12.

| Artist: Raphael | | | | | | | | |
|---|---|---|---|---|---|---|---|---|
| Painting Name | | CR | M1 | M2 | M3 | M | EMEC | EMEC2 |
| *'Allegory (The Knight's Dream)'* | fig. 7 | 1.50 | 2.35 | 2.58 | 4.62 | 3.03 | 23.08 | 35.42 |
| *'Madonna and Child'* | fig. 9 | 1.64 | 1.51 | 1.22 | 4.86 | 2.08 | 19.89 | 28.91 |
| *'Madonna Estergazi'* | | 1.54 | 2.11 | 2.43 | 5.71 | 3.08 | 24.36 | 32.07 |
| *'Canigiani Holy Family'* | fig. 8 | 1.50 | 1.70 | 1.46 | 3.48 | 2.05 | 23.97 | 27.74 |
| *'Colonna Madonna'* | fig. 10 | 1.49 | 2.97 | 1.63 | 7.89 | 3.37 | 20.46 | 20.98 |
| *'Madonna and Child Enthroned with Saints'* | | 1.38 | 1.12 | 1.11 | 4.18 | 1.73 | 16.39 | 38.68 |
| *'St. Catherine'* | fig. 12 | 1.51 | 1.06 | 1.09 | 7.27 | 2.03 | 25.14 | 24.09 |
| *'Crucifixion'* | | 1.41 | 1.61 | 1.65 | 2.29 | 1.83 | 16.91 | 28.21 |
| *'The Niccolini-Cowper Madonna'* | fig. 11 | 1.45 | 2.20 | 1.59 | 4.04 | 2.42 | 25.52 | 21.03 |
| …… | | ... | ... | ... | ... | ... | ... | ... |
| Average (over 30 paintings) | | 1.61 | 1.59 | 1.65 | 6.39 | 2.44 | 22.31 | 31.66 |

Table 10. Data for Raphael's paintings

| Artist: Leonardo da Vinci | | CR | M1 | M2 | M3 | M | EMEC | EMEC2 |
|---|---|---|---|---|---|---|---|---|
| Painting Name | | | | | | | | |
| *'Madonna Litta'* | fig. 5 | 1.56 | 1.53 | 1.36 | 5.24 | 2.22 | 25.72 | 29.98 |
| *'Virgin and Child with St. Anne'* | fig. 6 | 1.46 | 1.60 | 1.24 | 4.39 | 2.06 | 26.87 | 24.68 |
| *'Madonna Benois'* | | 1.55 | 1.71 | 2.18 | 5.86 | 2.80 | 29.62 | 32.44 |
| *'Madonna with the Carnation'* | | 1.57 | 2.45 | 2.46 | 4.47 | 3.00 | 23.02 | 20.42 |
| *'St. Hieronymus'* | | 1.61 | 3.64 | 3.86 | 4.90 | 4.10 | 26.49 | 37.01 |
| *'Madonna of the Rocks'* | | 1.48 | 0.79 | 0.96 | 6.78 | 1.73 | 24.46 | 30.95 |
| *'Portrait of an Unknown Woman (La Belle Ferroniere)'* | | 1.53 | 6.63 | 5.70 | 1.58 | 3.91 | 17.86 | 23.31 |
| *'Madonna of the Yarnwinder'* | | 1.48 | 1.18 | 1.08 | 6.43 | 2.02 | 22.43 | 29.86 |
| *'Portrait of Cecilia Gallerani (Lady with an Ermine)'* | | 1.60 | 4.95 | 6.17 | 2.15 | 4.03 | 24.20 | 30.53 |
| …… | | ... | ... | ... | ... | ... | ... | ... |
| Average (over 25 paintings) | | 1.46 | 2.86 | 2.84 | 5.84 | 2.67 | 24.35 | 26.35 |

Table 11. Data for Leonardo Da Vinci's paintings

| Artist: Rembrandt | | CR | M1 | M2 | M3 | M | EMEC | EMEC2 |
|---|---|---|---|---|---|---|---|---|
| Painting Name | | | | | | | | |
| *'Tobit and Anna'* | figs. 2,4 | 1.58 | 1.08 | 1.05 | 3.68 | 1.61 | 27.56 | 35.74 |
| *'Two Scholars Disputing (Peter and Paul?)'* | | 1.58 | 2.32 | 2.64 | 1.57 | 2.13 | 21.26 | 28.33 |
| *'Self-Portrait'* | | 1.67 | 2.31 | 2.20 | 2.31 | 2.28 | 21.16 | 25.08 |
| *'Christ in the Storm on the Lake of Galilee'* | | 1.46 | 1.58 | 1.52 | 1.30 | 1.46 | 17.61 | 21.04 |
| *'The Prophet Jeremiah Mourning over the Destruction of Jerusalem'* | | 1.65 | 2.03 | 2.35 | 3.75 | 2.61 | 24.94 | 32.28 |
| *'Philosopher Reading'* | | 1.44 | 2.86 | 3.24 | 3.57 | 3.21 | 29.85 | 23.25 |
| *'Portrait of a Young Woman with the Fan'* | | 1.42 | 2.27 | 2.44 | 1.80 | 2.15 | 20.19 | 23.07 |
| *'Feast of Belshazzar'* | | 1.49 | 1.65 | 1.80 | 4.19 | 2.32 | 21.40 | 33.48 |
| …… | | ... | ... | ... | ... | ... | ... | ... |
| Average (over 50 paintings) | | 1.65 | 2.71 | 2.96 | 3.13 | 2.70 | 22.01 | 26.65 |

Table 12. Data for Rembrandt's paintings

## Conclusion

In this work, we propose the model of prediction of color palletes in images of Renaissance oil artworks, by using the methods of enhancements, analyzing the color ratios and introducing the new measures for ratios of the primary colors in the RGB color model. Analyses of the data for each artist shows that the color ratio (CR) and the proposed measures (M) together with image enhancement measures can be considered as parameters to improve the quality of the existing image and predict the originality of colors in the artworks. The paintings of 105 artworks were used of the well-known artists, Rafael, Leonardo Da Vinci, and Rembrandt. All images of these paintings were downloaded from the Olga's Gallery by address http://www.abcgallery.com/. Analysis of paintings of other great artists of Renaissance and study in increasing the quality of the prediction accuracy will be the next step. In the future work, we are planning to use richer database reported in http://imag.pub.ro/pandora/pandora download.htm